\documentclass[11pt,a4paper]{article}
\usepackage[hyperref]{emnlp-ijcnlp-2019}
\usepackage{times}
\usepackage{latexsym}

\usepackage{url}

\usepackage{graphicx}  
\usepackage{amsmath}
\usepackage{amsfonts}
\usepackage{color}
\usepackage{multirow} 
\usepackage{booktabs,tabularx}       
\usepackage{makecell}
\definecolor{mygreen}{RGB}{28,172,0} 
\definecolor{mylilas}{RGB}{170,55,241}

\aclfinalcopy 

\title{Adversarial Reprogramming of Text Classification Neural Networks}

\author{Paarth Neekhara$^1$, Shehzeen Hussain$^2$, Shlomo Dubnov$^{1,3}$, Farinaz Koushanfar$^2$\\
$^1$Department of Computer Science\\
$^2$Department of Electrical and Computer Engineering\\
$^3$Department of Music\\
University of California San Diego\\
{\tt\footnotesize \{pneekhar,ssh028\}@ucsd.edu} \\
}

\date{}

\begin{document}

\maketitle
\begin{abstract}
In this work, we develop methods to repurpose text classification neural networks for alternate tasks without modifying the network architecture or parameters. We propose a context based vocabulary remapping method that performs a computationally inexpensive input transformation to reprogram a victim classification model for a new set of sequences. We propose algorithms for training such an input transformation in both white box and black box settings where the adversary may or may not have access to the victim model's architecture and parameters. We demonstrate the application of our model and the vulnerability of neural networks by adversarially repurposing various text-classification models including LSTM, bi-directional LSTM and CNN for alternate classification tasks.
\end{abstract}

\section{Introduction}

While neural network based machine learning models serve as the backbone of many text and image processing systems, recent studies have shown that they are vulnerable to adversarial examples. 
Traditionally, an adversarial example is a sample from the classifier's input domain which has been perturbed in such a way that is intended to cause a machine learning model to misclassify it. While the perturbation is usually imperceptible, such an adversarial input results in the neural network model outputting an incorrect class label with higher confidence. Several studies have shown such adversarial attacks to be successful in both the continuous input domain ~\cite{intriguing,limitations,blackbox,transferibility,advpatch,harnessing} and discrete input spaces ~\cite{discrete2,discrete3,discrete1}. 

Adversarial Reprogramming ~\cite{reprogramming} is a new class of adversarial attacks where a machine learning model is repurposed to perform a new task chosen by the attacker. The proposed attack is interesting because it allows an adversary to move a step beyond mere mis-classification of a victim network's output onto having the control to repurpose the network fully. The authors demonstrated how an adversary may repurpose a pre-trained ImageNet \cite{imagenet_cvpr09} model for an alternate classification task like classification of MNIST digits or CIFAR-10 images without modifying the network parameters. Since machine learning agents can be reprogrammed to perform unwanted actions as desired by the adversary, such an attack can lead to theft of computational resources such as cloud-hosted machine learning models. Besides theft of computational resources, the adversary may perform a task that violates the code of ethics of the system provider. 

The adversarial reprogramming approach proposed by \cite{reprogramming} trains an additive contribution $\theta$ to the inputs of the neural network to repurpose it for the desired alternate task. 
This approach assumes a white-box attack scenario where the adversary has full access to the network's parameters. Also, the adversarial program proposed in this work is only applicable to tasks where the input space of the the original and adversarial task is continuous.


In our work, we propose a method to adversarially
\mbox{\textit{repurpose}} neural networks which operate on sequences from a \textit{discrete} input space. The task is to learn a simple transformation (adversarial program) from the input space of the adversarial task to the input space of the neural network such that the neural network can be repurposed for the adversarial task. We propose a context-based vocabulary remapping function as an adversarial program for sequence classification networks. We propose training procedures for this adversarial program in both white-box and black-box scenarios. In the white-box attack scenario, where the adversary has access to the classifier's parameters, a Gumbel-Softmax trick \cite{gumbel_orig} is used to train the adversarial program. Assuming a black-box attack scenario, where the adversary may not have access to the classifier's parameters, we present a \mbox{REINFORCE} \cite{reinforceCITE} based optimization algorithm to train the adversarial program.
\begin{figure}[htbp]
\centering
\includegraphics[width=0.4\textwidth]{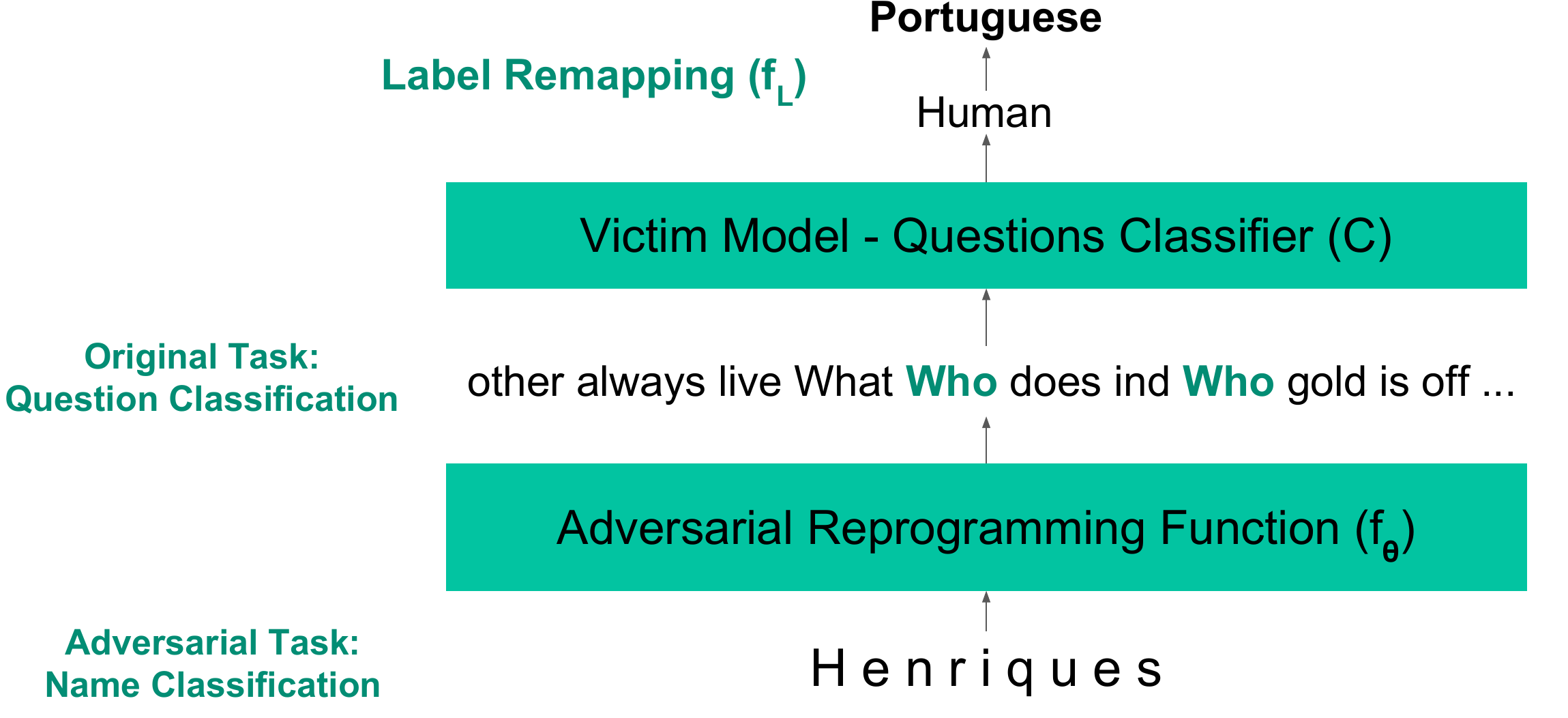}{\centering}
\caption{Example of Adversarial Reprogramming for Sequence Classification. We aim to design and train the adversarial reprogramming function $f_\theta$, such that it can be used to repurpose a pre-trained classifier C, for a desired adversarial task.}
\vspace{-0.3cm}
  \label{fig:frontimage}
\end{figure}

We apply our proposed methodology on various text classification models including Recurrent Neural Networks such as LSTMs and bidirectional LSTMs, and Convolutional Neural Networks (CNNs). We demonstrate experimentally, how these neural networks trained on a particular (original) text classification task can be repurposed for alternate (adversarial) classification tasks. We experiment with different text classification datasets given in table \ref{tab:dataset} as candidate original and adversarial tasks and adversarially reprogram the aforementioned text classification models to study the robustness of the attack.



\section{Background and Related Work}

\begin{figure*}[htbp]
\centering
\includegraphics[width=1.0\textwidth]{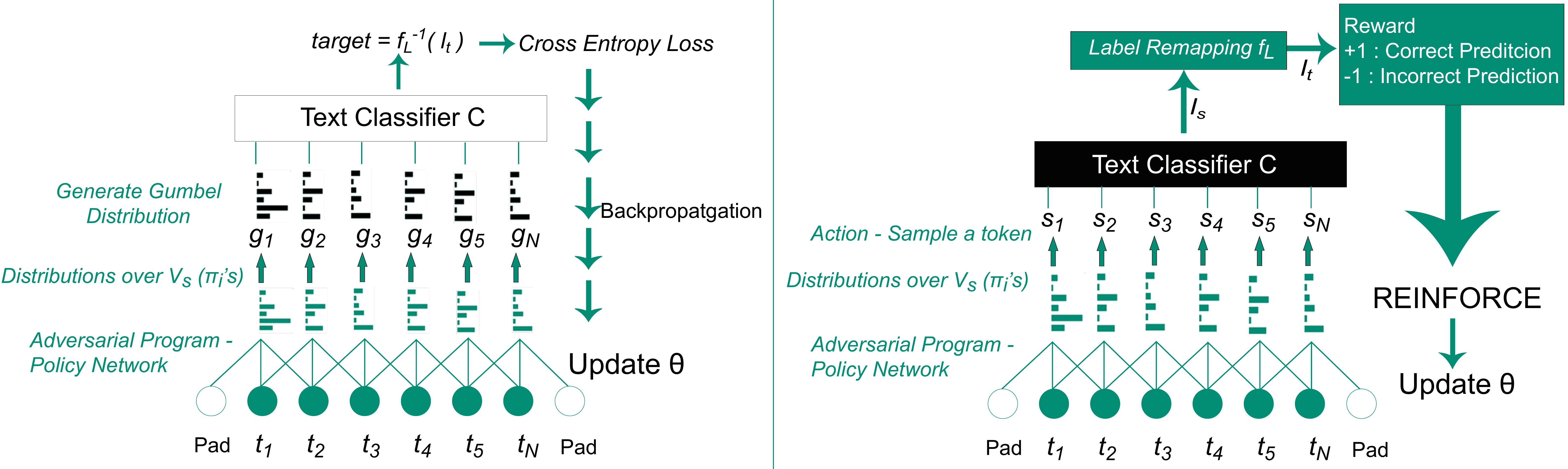}{\centering}
\caption{\textbf{Adversarial Reprogramming Function and Training Procedures.} \textbf{Left:} \textit{White-box} Adversarial Reprogramming. The adversary generates gumbel distributions $g_i$ at each time-step which are passed as a soft version of one-hot vectors to the classifier C. The cross-entropy loss between the predictions and the mapped class is backpropagated to train the adversarial program $\theta$. \textbf{Right:} \textit{Black-box} Adversarial Reprogramming. The adversarial reprogramming function is used as a policy network and the sampled action (sequence $s$) is passed to the classifier $C$ to get a reward based on prediction correctness. The adversarial program is then trained using REINFORCE.}
\vspace{-0.3cm}
  \label{fig:network}
\end{figure*}

\subsection{Adversarial Examples}
Traditionally, adversarial examples are intentionally designed inputs to a machine learning model that cause the model to make a mistake \cite{harnessing}. These attacks can be broadly classified into \textit{untargeted} and \textit{targeted} attacks. In the untargeted attack scenario, the adversary succeeds as long as the victim model classifies the adversarial input into \textit{any} class other than the correct class, while in the \textit{targeted} attack scenario, the adversary succeeds only if the model classifies the adversarial input into a specific incorrect class. In both these scenarios, the intent of the adversary is usually malicious and the outcome of the victim model is still limited to the original task being performed by the model.

Adversarial attacks of image-classification models often use gradient descent on an image to create a small perturbation that causes the machine learning model to mis-classify it \cite{intriguing,evasion}. 
There has been a similar line of adversarial attacks on neural networks with discrete input domains \cite{discrete2,discrete1}, where the adversary modifies a few tokens in the input sequence to cause mis-classification by a sequence model. In addition, efforts have been made in designing more general adversarial attacks in which the same modification can be applied to many different inputs to generate adversarial examples \cite{advpatch,harnessing,universal}. For example,  authors \cite{ATN} trained an Adversarial Transformation Network that can be applied to all inputs to generate adversarial examples targeting a victim model or a set of victim models. In this work, we aim to learn such universal transformations of discrete sequences for a fundamentally different task: \textit{Adversarial Reprogramming} described below.

\subsection{Adversarial Reprogramming}
Adversarial Reprogramming \cite{reprogramming} introduced a new class of adversarial attacks where the adversary wishes to repurpose an existing neural network for a new task chosen by the attacker, without the attacker needing to compute the specific desired output. The adversary achieves this by first defining a hard-coded one-to-one label remapping function $h_g$ that maps the output labels of the adversarial task to the label space of the classifier $f$; and learning a corresponding adversarial reprogramming function $h_f(.;\theta)$ that transforms an input $(\tilde{X})$ \footnote{$\tilde{X}$ is an ImageNet size $(n\times n\times 3)$ padded input image} from the input space of the new task to the input space of the classifier. The authors proposed an adversarial reprogramming function $h_f(.;\theta)$, for repurposing ImageNet models for adversarial classification tasks. An adversarial example $X_{adv}$ for an input image $\tilde{X}$ can be generated using the following adversarial program: \footnote{Masking ignored because it is only a visualization convenience}
$$X_{adv} = h_f(\tilde{X};\theta) = \tilde{X} + \tanh(\theta)$$
where $\theta \in \mathbb{R}^{n \times n \times 3}$ is the learnable weight matrix of the adversarial program (where n is the ImageNet image width).  Let $P(y | X)$ denote the probability of the victim model predicting label $y$ for an input $X$. The goal of the adversary is to maximize the probability $P(h_g(y_{adv}) | X_{adv})$ where $y_{adv}$ is the label of the adversarial input $X_{adv}$. The following optimization problem that maximizes the log-likelihood of predictions for the adversarial classification task, can be solved using backpropagation to train the adversarial program parameterized by $\theta$:
\begin{equation}
\hat{\theta} = 
argmin_{\theta} \left( 
	- \log P(h_g(y_{adv}) | X_{adv}) + \lambda ||\theta||_2^2
	\right)
\label{eq:repr}
\end{equation}
where $\lambda$ is the regularization hyperparameter. Since the adversarial program proposed is a trainable additive contribution $\theta$ to the inputs, it's application is limited to neural networks with a continuous input space. Also, since the the above optimization problem is solved by back-propagating through the victim network, it assumes a white-box attack scenario where the adversary has gained access to the victim model's parameters. In this work, we describe how we can learn a simple transformation in the \textit{discrete space} to extend the application of adversarial reprogramming on \textit{sequence classification} problems. We also propose a training algorithm in the black-box setting where the adversary may not have access to the model parameters.

\subsection{Transfer Learning}
Transfer Learning ~\cite{Raina07self-taughtlearning:} is a study closely related to Adversarial Reprogramming. During training, neural networks learn representations that are generic and can be useful for many related tasks. A pre-trained neural network can be effectively used as a feature extractor and the parameters of just the last few layers are retrained to realign the output layer of the neural network for the new task. Prior works have also applied transfer learning on text classification tasks \cite{transferlearning,texttransfer}. While transfer learning approaches 
exploit the learnt representations for the new task, they cannot be used to repurpose an exposed neural network for a new task without modifying some intermediate layers and parameters.

Adversarial Reprogramming \shortcite{reprogramming} studied whether it is possible to keep all the parameters of the neural network unchanged and simply learn an input transformation that can realign the outputs of the neural network for the new task. This makes it possible to repurpose \textit{exposed} neural network models like cloud-based photo services to a new task where transfer learning is not applicable since we do not have access to intermediate layer outputs. 

\section{Methodology}

\subsection{Threat Model}

Consider a sequence classifier $C$ trained on the original task of mapping a sequence $s \in S$ to a class label $l_S \in L_S$ i.e $C : s \mapsto l_S $. 
An adversary wishes to repurpose the original classifier $C$ for the adversarial task $C'$ of mapping a sequence $t \in T$ to a class label $l_T \in L_T$ i.e $C' : t \mapsto l_T $.
The adversary can achieve this by hard-coding a one-to-one label remapping function: $$f_L :  l_S \mapsto l_T$$ that maps an original task label to the new task label and learning a corresponding adversarial reprogramming function: $$f_\theta : t \mapsto s$$ that transforms an input from the input space of the adversarial task to the input space of the original task. The adversary aims to update the parameters $\theta$ of the adversarial program $f_\theta$ such that the mapping $f_L( C ( f_{\theta} (t) ) )$ can perform the adversarial classification task $C' : t \mapsto l_T$.

\subsection{Adversarial Reprogramming Function}
\label{sec:rf}
The goal of the adversarial reprogramming function $f_\theta : t \mapsto s$ is to map a sequence $t$ to $s$ such that it is labeled correctly by the classifier $f_L (C )$. 

The tokens in the sequence $s$ and $t$ belong to some vocabulary lists $V_S$ and $V_T$ respectively. We can represent the sequence $s$ as $s = s_1,s_2,..,s_N$ where $s_i$ is the vocabulary index of the $i_{th}$ token in sequence $s$ in the vocabulary list $V_S$.
Similarly sequence $t$ can be represented as $t = t_1,t_2,..,t_N$ where $t_i$ is the vocabulary index of the $i_{th}$ token of sequence $t$ in the vocabulary list $V_T$.

In the simplest scenario, the adversary may try to learn a vocabulary mapping from $V_T$ to $V_S$ using which each $t_i$ can be independently mapped to some $s_i$ to generate the adversarial sequence. Such an adversarial program has limited potential since the representational capacity of such a reprogramming function is very limited. We experimentally support this hypothesis by showing how such a transformation has limited potential for the purpose of adversarial reprogramming.

A more sophisticated adversarial program can be a sequence to sequence machine translation model \cite{seq2seq} that learns a translation $t \mapsto s$ for adversarial reprogramming. While theoretically this is a good choice, it defeats the purpose of adversarial reprogramming. This is because the computational complexity of training and using such a machine translation model would be similar if not greater than that of a new sequence classifier for the adversarial task $C'$.

The adversarial reprogramming function should be computationally inexpensive but powerful enough for adversarial repurposing. To this end, we propose a context-based vocabulary remapping model that produces a distribution over the target vocabulary at each time-step based on the surrounding input tokens. More specifically, we define our adversarial program as a trainable 3-d matrix $\theta_{k \times |V_T| \times |V_S| } $ where k is the context size. Using this, we generate a probability distribution $\pi_i$ over the vocabulary $V_S$ at each time-step $i$ as follows:
\begin{equation} \label{hi}
h_i = \sum_{j=0}^{k-1} \theta[j,t_{i + \lfloor k/2 \rfloor - j}]
\end{equation}
\begin{equation} \label{pi}
\pi_i = softmax(h_i)
\end{equation}
Both $h_i$ and $\pi_i$ are vectors of length $|V_S|$. To generate the adversarial sequence $s$ we sample each $s_i$ independently from the distribution $\pi_i$ i.e $s_i \sim \pi_i$


In practice, during training, we implement this adversarial program as a single layer of 1-d convolution over the sequence of one-hot encoded vectors of adversarial tokens $t_i$ with $|V_T|$ input channels and $|V_S|$ output channels with $k$-length kernels parameterized by $\theta_{k \times |V_T| \times |V_S| }$. 



\subsection{White-box Attack}
In the white-box attack scenario, we assume that the adversary has gained access to the victim network's parameters and architecture. To train the adversarial reprogramming function $f_{\theta}$, we use an optimization objective similar to equation \ref{eq:repr}. Let $P(l | s)$ denote the probability of predicting label $l$ for a sequence $s$ by classifier $C$. We wish to maximize the probability $P( f_L^{-1}(l_t) | f_{\theta}( t) )$ which is the probability of the output label of the classifier being mapped to the correct class $l_t$ for an input $t$ in the domain of the adversarial task. Therefore we need to solve the following log-likelihood maximization problem:

\begin{equation}
\hat{\theta} = argmin_{\theta}(-\sum_t log(P( f_L^{-1}(l_t) | f_{\theta}( t))) )
\label{eq:objective}
\end{equation}

Note that that the output of the adversarial program $s = f_{\theta}(t)$ is a sequence of discrete tokens. This makes the above optimization problem non-differentiable. Prior works \cite{gumbelgan,nmtgumbel,discrete1} have demonstrated how we can smoothen such an optimization problem using the Gumbel-Softmax \cite{gumbel_orig} distribution.

In order to backpropagate the gradient information from the classifier to the adversarial program, we smoothen the generated tokens $s_i$ using Gumbel-Softmax trick as per the following:

For an input sequence $t$, we generate a sequence of Gumbel distributions $g = g_1,g_2,..,g_N$. The $n_{th}$ component of distribution $g_i$ is generated as follows:
$$g_i^n =  \frac{\exp ( (\log(\pi_{i}^{n}) + r_n)/temp)}{\sum_j \exp ( (\log(\pi_{i}^{j}) + r_j)/temp)}$$

where $\pi_i$ is the softmax distribution at the $i_{th}$ time-step obtained using equation \ref{pi}, $r_n$ is a random number sampled from the Gumbel distribution \cite{gumbel1954statistical} and $temp$ is the temperature of Gumbel-Softmax.

The sequence then passed to the classifier $C$ is the sequence $g$ which serves as a soft version of the one-hot encoded vectors of $s_i$'s. Since the model is now differentiable, we can solve the following optimization problem using backpropagation: 
$$
\hat{\theta} = argmin_{\theta}(-\sum_t log(P( f_L^{-1}(l_t) | g)) )
$$


During training the temperature parameter is annealed from some high value $t_{max}$ to a very low value $t_{min}$. The details of this annealing process for our experiments have been included in the appendix.



\subsection{Black-box Attack}
In the black-box attack scenario, the adversary can only query the victim classifier $C$ for labels (and not the class probabilities). We assume that the adversary has the knowledge of the vocabulary $V_S$ of the victim model. Since the adversarial program needs to produce a discrete output to feed as input to the classifier $C$, it is not possible to pass the gradient update from the classifier $f_L(C)$ to the adversarial program $\theta$ using standard back-propagation. Also, in the black-box attack setting it is not possible to back-propagate the cross entropy loss through the classifier $C$ in the first place.

We formulate the sequence generation problem as a Reinforcement Learning problem \cite{datagen,actorcritic,seqgan} where the adversarial reprogramming function is the policy network. The state of the adversarial program is a sequence $t \in T$ and an action of our policy network is to produce a sequence of tokens $s \in S$. The adversarial program parameterized by $\theta$, models the stochastic policy $\pi_{adv}(s | t; \theta)$ such that a sequence $s \in S$ may be sampled from this policy conditioned on $t \in T$. We use a simple reward function where we assign a reward +1 for a correct prediction and -1 for an incorrect prediction using the classifier $f_L(C)$ where $f_L$ is the label remapping function and $C$ is the classifier.
Formally:
\[
    r(t,s)= 
\begin{cases}
    +1,& f_L(C(s)) = l_t \\
    -1,& f_L(C(s)) \neq l_t
\end{cases}
\]
The optimization objective to train the policy network is the following:
\begin{eqnarray}
\max_{\theta} 
J(\theta) \quad \text{where,} \quad \nonumber
J(\theta) = \mathbb{E}_{\pi_{adv}}[r(t, s)] \nonumber
\end{eqnarray}
Following the REINFORCE algorithm \cite{reinforceCITE} we can write the gradient of the expectation with respect to $\theta$ as per the following:

\begin{eqnarray}
\nabla_{\theta}J 
&=& \nabla_{\theta} \left[\mathop{\mathbb{E}}_{\pi_{adv}} \left[ r(t, s) \right] \right] \nonumber \\
&=& \nabla_{\theta} \left[ \sum_{s}\pi_{adv}(s|t;\theta)r(t, s)  \right] \nonumber \\
&=& \sum_{s}\pi_{adv}(s|t;\theta) \nabla_{\theta} \log(\pi_{adv}(s|t;\theta)) r(t, s) \nonumber \\
&=& \mathop{\mathbb{E}}_{\pi_{adv}}\left[ r(t, s) \, \nabla_{\theta}\log(\pi_{adv}(s|t;\theta))  \right] \nonumber\\
&=& \mathop{\mathbb{E}}_{\pi_{adv}}\left[ r(t, s) \, \nabla_{\theta}\log(\pi_{adv}(s_1, .., s_N|t;\theta))  \right] \nonumber\\
&=& \mathop{\mathbb{E}}_{\pi_{adv}}\left[ r(t, s) \, \nabla_{\theta}  \log (\prod_{i} \pi_{adv}(s_i|t;\theta))  \right] \nonumber\\
&=& \mathop{\mathbb{E}}_{\pi_{adv}}\left[ r(t, s) \, \sum_{i} \nabla_{\theta} \log(\pi_{adv}(s_i|t;\theta))  \right] \nonumber
\end{eqnarray}

Note that $\pi_{adv}(s_i|t;\theta)$ is the same as $\pi_i$ defined in equation \ref{pi} which can be differentiated with respect to $\theta$. The expectations are estimated as sample averages. Having obtained the gradient of expected reward, we can use mini-batch gradient ascent to  update $\theta$ with a learning rate $\alpha$ as: $\theta \leftarrow \theta + \alpha \nabla_{\theta}J$.

\section{Experiments}

\subsection{Datasets and Classifiers}

We demonstrate the application of the proposed reprogramming techniques on various text-classification tasks. In our experiments, we design adversarial programs to attack both word-level and character-level text classifiers. Additionally, we aim to adversarially repurpose a character-level text classifier for a word-level classification task and vice-versa. To this end, we choose the following text-classification datasets as candidates for the original and adversarial classification tasks:

\textbf{\textit{1. Surname Classification Dataset} (\textbf{Names-18, Names-5})~\cite{Names}}: The dataset categorizes surnames from 18 languages of origin.  
We use a subset of this dataset \textit{Names-5} containing Names from 5 classes: \textit{Dutch, Scottish, Polish, Korean} and \textit{Portuguese}, as a candidate for adversarial task in the experiments. \newline\textbf{\textit{2. Experimental Data for Question Classification} (\textbf{Questions})~\cite{TREC}}: categorizes around 5500 questions into 6 classes: \textit{Abbreviation, Entity, Description, Human, Location, Numeric}. 
\newline\textbf{\textit{3. Sentiment Classification Dataset~\cite{Arabic}} (Arabic) }: contains 2000 binary labeled tweets on diverse topics such as politics and arts. 
\newline\textbf{\textit{4. Large Movie Review Dataset} \textbf{(IMDB)} for sentiment classification ~\cite{IMDB}}: contains 50,000 movie reviews categorized into binary class of positive and negative sentiment.

\vspace{-0.1mm}
The statistics of the above mentioned datasets have been given in table \ref{tab:dataset}. We train adversarial reprogramming functions to repurpose various text-classifiers based on Long Short-Term Memory (LSTM) network \cite{THE_LSTM_PAPER}, bidirectional LSTM network \cite{biLSTM} and Convolutional neural network \cite{kim2014convolutional} models. All the aforementioned models can be trained for both word-level and character-level classification. We use character level classifiers for \textit{Names-18} and \textit{Names-5} datasets and word-level classifiers for \textit{IMDB}, \textit{Questions} and \textit{Arabic} datasets. We use randomly initialized word/character embeddings which are trained along with the classification model parameters. For LSTM, we use the output at last timestep for prediction. For the Bi-LSTM, we combine the outputs of the first and last time step for prediction. For the Convolutional Neural Network we follow the same architecture as \cite{kim2014convolutional}. Additionally, to understand the effectiveness of adversarial reprogramming, we train a sigle layer CNN with convolutional filter width 5 and discuss the results of this experiment in Section \ref{sec:complexity}. The details of these classifiers have been included in table 2 of the supplementary material.

\setlength\tabcolsep{3pt} 

\begin{table}[bt!]
\resizebox{0.47\textwidth}{!}{
 \begin{tabular}{l|c|c|c|c|c}
 \toprule
 \thead{Data Set} & \thead{\# Classes} & \thead{Train\\Samples} & \thead{Test\\Samples} & \thead{$|V|$} & \thead{Avg \\Length}\\
\midrule
Names-18 & 18 & 115,028 & 28,758 & 90 & 7.1 \\ 
Names-5 & 5 & 3632 & 909 & 66 & 6.5 \\ 
Questions & 6 & 4361 & 1091 & 1205 & 11.2 \\ 
Arabic & 2 & 1600 & 400 & 955 & 9.7 \\ 
IMDB & 2 & 25,000 & 25,000 & 10000 & 246.8 \\ 
 \bottomrule
 \end{tabular} 
 }
 \vspace{-2mm}
 \caption{Summary of datasets. $|V|$ denotes the vocabulary size of each dataset.}
\label{tab:dataset}
\end{table}

\begin{table}[bt!]
\centering
\resizebox{0.37\textwidth}{!}{

 \begin{tabular}{l|ccc|c}
 \toprule
\thead{Data Set} &  \thead{LSTM} & \thead{Bi-LSTM} & \thead{CNN} & \thead{1-CNN}\\
\midrule
Names-18  & 97.84 & 97.84 & \textbf{97.88} & 74.04\\ 
Names-5  &\textbf{99.88} & \textbf{99.88} & 99.77 & 71.51\\ 
Questions & 96.70 & \textbf{98.25} & 98.07 & 83.77\\ 
Arabic & 87.25 &  \textbf{88.75}& 88.00 & 74.75\\ 
IMDB & 86.83 & 89.43 & \textbf{90.02} & 83.32\\ 
 \bottomrule
 \end{tabular} 
 } 
\vspace{-2mm}
 \caption{Test accuracy of various classification models. We use character-level models for \textit{Names-5} and \textit{Names-18} and word-level models for all other tasks. 1-CNN is a single layer CNN model with filter width 5. }
\label{tab:classifiers}
\end{table}

\subsection{Experimental Setup}
As described in the methodology section, the label remapping function $f_L$ we use, is a one-to-one mapping between the labels of the original task and the adversarial task. Therefore it is required to apply the constraint that the number of classes of the adversarial task are less than or equal to the number of classes of the original task. We choose  \textit{Names-5, Arabic} and \textit{Question Classification} as candidates for the adversarial tasks and repurpose the models allowed under this constraint. We use context size $k=5$ for all our experiments.

In white-box attacks, we use the Gumbel-Softmax based approach described in the methodology to train the adversarial program. The details of the temperature annealing process are included in table 1 of the supplementary material.
For black-box attacks, we use the REINFORCE algorithm described in methodology, on mini-batches of sequences. Since the action space for certain reprogramming problems, (eg. reprogramming of IMDB classifier) is large $(|V_S| = 10000)$, we restrict the output of the adversarial program to most frequent 1000 tokens in the vocabulary $V_S$. We use Adam optimizer \cite{adam} for all our experiments. Hyperparameter details of all our experiments are included in table 1 of the supplementary material.

\begin{table}[t]

  \begin{center}
    
    \resizebox{0.475\textwidth}{!}{
    \begin{tabular}{c|l|l|l|l|l}
    \toprule
    	\textbf{} & \textbf{} & \textbf{} & \multicolumn{3}{c}{\textbf{Test Accuracy (\%)}}\\
      \thead{Victim\\Model} & \thead{Original\\ Task} & \thead{Adversarial \\Task} & \thead{Black\\box} & \thead{White\\Box} & \thead{White\\Box on\\Random \\Network}\\
      \midrule
      \multirow{4}{*}{LSTM} 
      & Questions & Names-5 & 80.96 & 97.03 &  44.33 \\
      & Questions & Arabic & 73.50 & 87.50 &  50.00 \\
      & Names-18 & Questions & 68.56 & 95.23  & 28.23 \\
      & Names-18 & Arabic & 83.00 & 84.75 &  51.50 \\
      & IMDB& Arabic & 80.75  & \textbf{88.25} &  50.50\\     \midrule
      \multirow{4}{*}{Bi-LSTM} 
      & Questions & Names-5 & \textbf{93.51} & \textbf{99.66} &  63.14 \\
      & Questions & Arabic & 81.75 & 83.50 &  70.00 \\
      & Names-18 & Questions & \textbf{94.96} & 97.15 &  80.01 \\
      & Names-18 & Arabic & 78.75 & 84.25 &  69.25 \\
      & IMDB& Arabic & 83.25 & 86.75 &  84.00 \\
      \midrule
      \multirow{4}{*}{CNN} 
      & Questions & Names-5 & 88.90 & 99.22 &  93.06 \\
      & Questions & Arabic & 82.25  & 87.25 &  76.25 \\
      & Names-18 & Questions & 71.03 & \textbf{97.61} &  33.45 \\
      & Names-18 & Arabic & 80.75 & 86.50 &  60.00 \\
      & IMDB& Arabic & \textbf{84.00} & 87.00 &  84.25\\
      \midrule
    \end{tabular}
    }
  \end{center}
\vspace{-0.3cm}
\caption{\textbf{Adversarial Reprogramming Experiments:} The accuracies of \textit{white-box} and \textit{black-box} reprogramming experiments on different combinations of original task, adversarial task and model. Figures in bold correspond to our best results on a particular adversarial task in the given attack scenario (black-box and white-box). \textit{White-box on Random Network} column presents results of the white-box attack on an untrained neural network. Context size $k=5$ is used for all our experiments.}
\label{tab:table2}

\end{table}

\subsection{Results and Discussion}
The accuracies of all adversarial reprogramming experiments have been reported in table \ref{tab:table2}. To interpret the results in context, the accuracies achieved by the LSTM, Bi-LSTM and CNN text classification models on the adversarial tasks can be found in table \ref{tab:dataset}.

We demonstrate how character-level models trained on Names-18 dataset can be repurposed for word-level sequence classification tasks like Question Classification and Arabic Tweet Sentiment Classification. Similarly, word-level classifiers trained on Question Classification Dataset can be repurposed for the character-level Surname classification task. Interestingly, classifiers trained on IMDB Movie Review Dataset can be repurposed for Arabic Tweet Sentiment Classification even though there is a high difference between the vocabulary size (10000 vs 955) and average sequence length(246.8 vs 9.7) of the two tasks. It can be seen that all of the three classification models 
are susceptible to adversarial reprogramming in both white-box and black-box setting.

White-box based reprogramming outperforms the black-box based approach in all of our experiments. In practice, we find that training the adversarial program in the black-box scenario requires careful hyper-parameter tuning for REINFORCE to work. 
We believe that improved reinforcement learning techniques for sequence generation tasks \cite{actorcritic,datagen} can make the training procedure for black-box attack more stable. We propose such improvement as a direction of future research. 

To assess the importance of the original task on which the network was trained, we also present results of white-box adversarial reprogramming on untrained random network. Our results are coherent with similar experiments on adversarial reprogramming of untrained ImageNet models \cite{reprogramming} demonstrating that adversarial reprogramming is less effective when it targets untrained networks. The figures in table \ref{tab:table2} suggest that the representations learned by training a text classifier on an original task, are important for repurposing it for an alternate task. However another plausible reason as discussed by \citeauthor{reprogramming} is that the reduced performance on random networks might be because of simpler reasons like poor scaling of network weight initialization making the optimization problem harder.

\subsection{Complexity of Reprogramming Function}
\label{sec:complexity}
As motivated earlier in Section \ref{sec:rf}, computational efficiency of the adversarial program is critical for adversarial reprogramming to be of interest to an adversary. If the adversary can achieve the desired results using a computationally inexpensive classifier, it defeats the purpose of adversarial reprogramming. To understand if this is the case, we train a one-layer CNN with the same convolutional filter width as our adversarial program and average the activations across all time-steps to classify a given phrase. The results of such a classifier on various datasets have been reported in Table \ref{tab:classifiers}. We can observe that our white-box attack on pre-trained networks, outperforms this classifier in all scenarios (refer to Table \ref{tab:table2}). Our best black-box attacks also outperform a one-layer CNN for all adversarial tasks. This experiment demonstrates that the reprogramming function exploits the learned feature representation of the victim model. Also, the observation that adversarial reprogramming is significantly less effective on randomly initialized untrained networks further reinforces the importance of utilizing a trained victim model. 

Since the reprogramming function is a context based vocabulary remapping function, we can implement it as a look-up table that maps a combination of $k$ tokens from the vocabulary $V_t$ to a token in the source vocabulary $V_s$. The time complexity for transforming a sentence $t$ to an adversarial sentence is just $O(length(t))$.

\subsubsection{Adversarial Sequences} 
\begin{figure}[htp]
\includegraphics[width=0.474\textwidth]{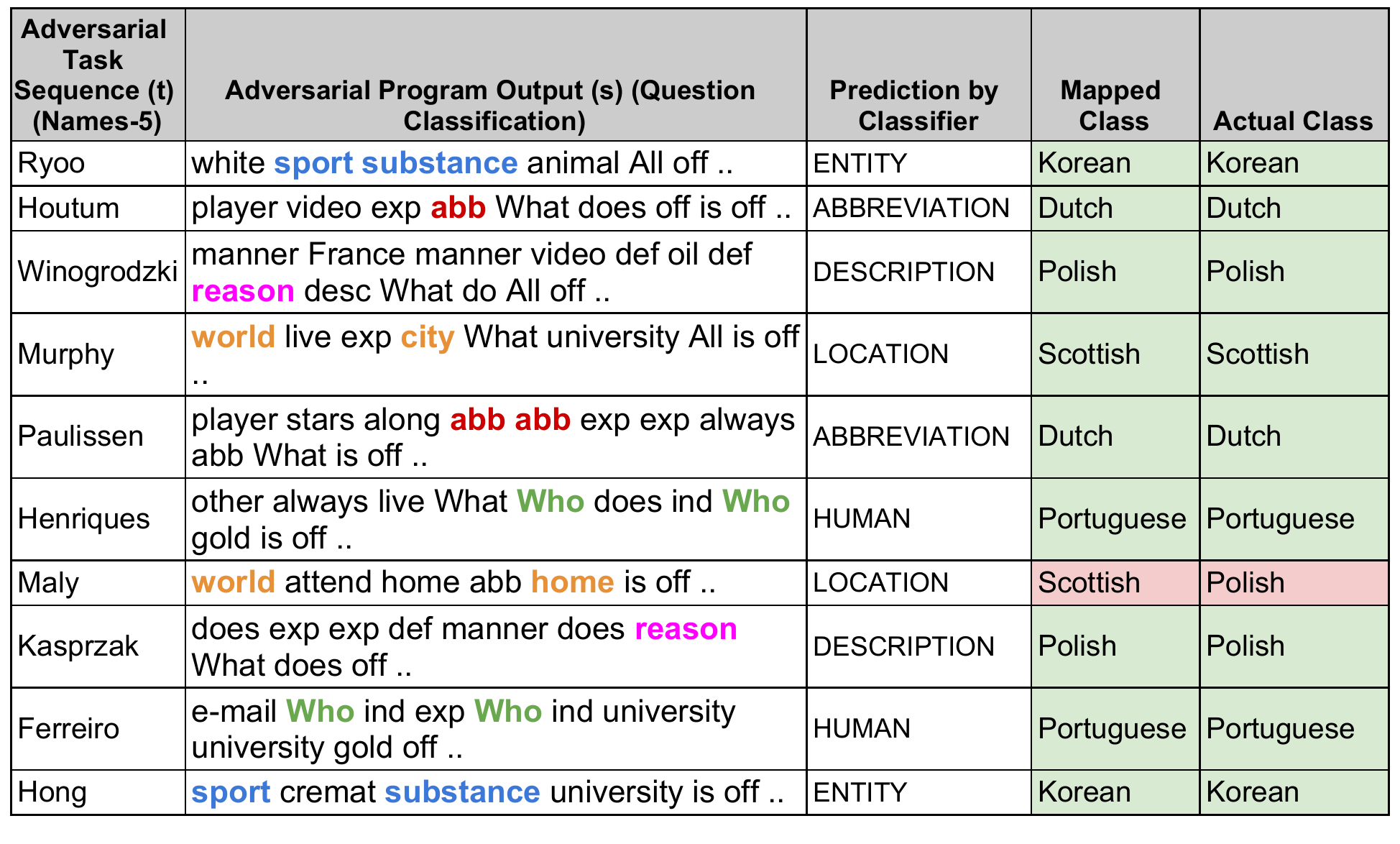}{\centering}
\caption{Adversarial sequences generated by our adversarial program for Names-5 Classification (adversarial task), when targeting a CNN trained on the Question Classification dataset (original task). Interpretable transformations are shown as colored words in the second column.
}


\label{fig:samples}
\end{figure}

Figure \ref{fig:samples} shows some adversarial sequences generated by the adversarial program for Names-5 Classification while attacking a CNN trained on the Question Classification dataset. A sequence $t$ in the first column is transformed into the adversarial sequence $s$ in the second column by the trained adversarial reprogramming function. 
While these adversarial sequences may not make semantic or grammatical sense, it exploits the learned representation of the classifier to map the inputs to the desired class. For example, sequences that should be mapped to \mbox{HUMAN} class have words like \textit{Who} in the generated adversarial sequence. Similarly, sequences that should be mapped to \mbox{LOCATION} class have words like \textit
{world, city} in the adversarial sequence. Other such interpretable transformations are depicted via colored text in the adversarial sequences of Figure \ref{fig:samples}.

\vspace{-2mm}
\subsubsection{Effect of Context Size}
\begin{figure}[htp]
\centering
\includegraphics[width=0.47\textwidth]{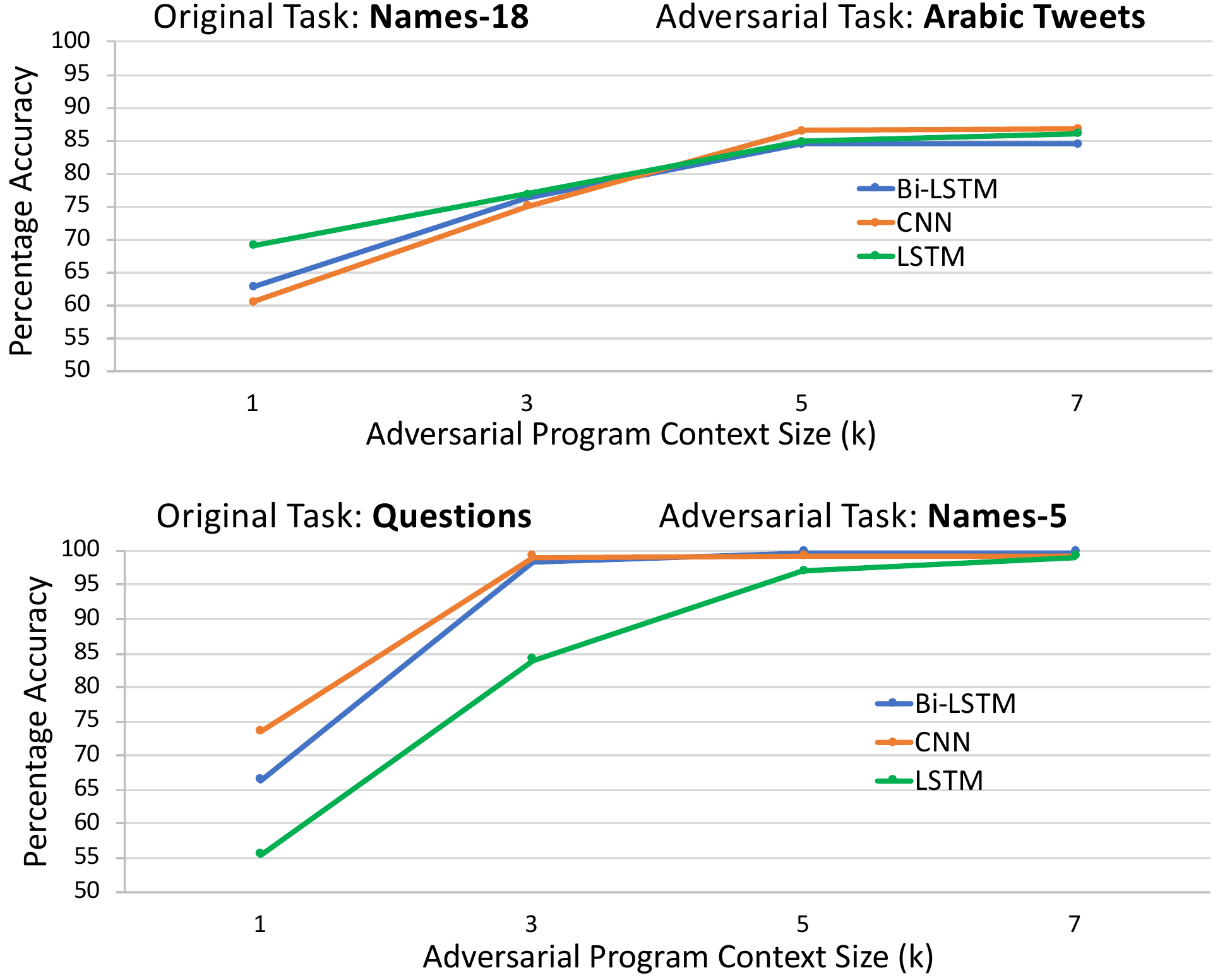}{\centering}
\vspace{-2mm}
\caption{Accuracy vs Context size ($k$) plots for all 3 classification models on 2 different adversarial reprogramming experiments.}
\vspace{-0.3cm}
  \label{fig:filterwidth}
\end{figure}

By varying the context size $k$ of the convolutional kernel $\theta_{k\times|V_T|\times|V_S|}$ in our adversarial program we are able to control the representational capacity of the adversarial reprogramming function. Figure \ref{fig:filterwidth} shows the percentage accuracy obtained when training the adversarial program with different context sizes $k$ on two different adversarial tasks: Arabic Tweets Classification and Name Classification. Using a context size $k = 1$ reduces the adversarial reprogramming function to simply a vocabulary remapping function from $V_S$ to $V_T$. It can be observed that the performance of the adversarial reprogramming model at $k=1$ is significantly worse than that at higher values of $k$. While higher values of $k$ improve the performance of the adversarial program, they come at a cost of increased computational complexity and memory required for the adversarial reprogramming function. For the adversarial tasks studied in this paper, we observe that $k=5$ is a reasonable choice for context size of the adversarial program.

\vspace{-0.5em}
\section{Conclusion}
In this paper, we extend adversarial reprogramming, a new class of adversarial attacks, to target text classification neural networks. Our results demonstrate the effectiveness of such attacks in the more challenging black-box settings, posing them as a strong threat in real-world attack scenarios. 
We demonstrate that neural networks can be effectively reprogrammed for alternate tasks, which were not originally intended by a service provider. Our proposed end-to-end approach can be used to further understand the vulnerabilities and blind spots of deep neural network based text classification systems. Furthermore, due to the threat presented by adversarial reprogramming, we recommend future work to study defenses against such attacks.

\bibliography{emnlp-ijcnlp-2019}
\bibliographystyle{acl_natbib}

\end{document}